%% file: 0.tex
\documentclass[11pt]{stacs}
 
\usepackage{color}
\usepackage{graphicx}
\usepackage{epsfig}
\usepackage{cite}
\usepackage{amsmath, amsxtra, amsfonts}%
\usepackage{eucal}
\usepackage{xspace, varioref}
\usepackage{mdwtab}
\usepackage{dcolumn}
\usepackage{xspace}
\date{}

\input local-defs.tex

\newcommand{\X}{ll\xspace}

\begin{document}

\title{A Fast Algorithm and Datalog Inexpressibility for Temporal Reasoning}
\author{M. Bodirsky}{Manuel Bodirsky}
\address{Humboldt-Universit\"at zu Berlin, Germany}
\email{bodirsky@informatik.hu-berlin.de}
\author{J. K\'ara}{Jan K\'ara}
\address{Charles University, Prague, Czech Republic}
\email{kara@kam.mff.cuni.cz}
\thanks{Supported by Project 1M0021620808 of the Ministry of Education of the Czech Republic.}
\input abstract.tex
\maketitle
\input introduction.tex
\input tcsps.tex

\input algebraic.tex

\input algorithm.tex

\input unbounded.tex
\input outlook.tex

\section*{Acknowledgements}
We would like to thank Christopher Rudolf for implementing the algorithm, and the anonymous referees for their helpful comments.
\bibliographystyle{abbrv}
\bibliography{../../local}
\end{document}

%% file: local-defs.tex
\newcommand{\cproblem}[3]{
\vspace{.2cm}
\noindent {\bf #1} \\
INSTANCE: #2 \\
QUESTION: #3 \\}

\newcommand{\cal}[1]{\ensuremath{\mathcal{#1}}}
\newcommand{\lex}{\ensuremath{\text{lex}}}
\newcommand{\lele}{\ensuremath{\text{ll}}}

\newcommand{\true}{\ensuremath{\mathtt{true}}}
\newcommand{\false}{\ensuremath{\mathtt{false}}}

\newcommand{\ignore}[1]{}

\newcommand{\tcl}{TCL}
\newcommand{\tcls}{TCLs}


%% file: abstract.tex
\begin{abstract}
We introduce a new tractable temporal constraint language, which
strictly contains the Ord-Horn language of B\"urkert and Nebel and the class of AND/OR precedence constraints. The algorithm we
present for this language decides whether a given set of constraints is
consistent in time that is quadratic in the input size.  We also prove that
(unlike Ord-Horn) the constraint satisfaction problem of this language cannot be solved by Datalog or by
establishing local consistency.
\end{abstract}
\keywords{constraint satisfaction, temporal reasoning, computational complexity, Ord-Horn, algorithms, Datalog, precedence constraints}

%% file: introduction.tex
\section{Introduction}
Temporal reasoning plays an important role in Artificial Intelligence. Almost
any area in AI -- for instance common-sense reasoning, natural language
processing, scheduling, planning -- involves some sort of temporal reasoning.
In 1993, Golumbic and Shamir~\cite{GolumbicShamir} listed applications of
temporal reasoning problems in archeology, behavioral psychology, operations
research, and circuit design.  Since then, temporal reasoning became one of the benchmark applications of
constraint processing in general~\cite{DechterBook}.  Contributions to the
field have various background, for example database theory~\cite{vanderMeyden}, scheduling~\cite{and-or-scheduling},
constraint satisfaction complexity~\cite{CSPSurvey}, the theory of relation
algebras~\cite{LadkinMaddux,Duentsch}, combinatorics~\cite{GolumbicShamir}, and
artificial intelligence~\cite{HandbookTR}.

This paper deals with \emph{temporal constraint languages}.
A \emph{temporal constraint language (\tcl)} is a countable collection
of relations with a first-order definition in $(\mathbb Q,<)$, the 
linear order of the rational numbers; a
detailed definition is given in Section~\ref{sect:tcsps}.
One of the most fundamental TCLs 
is the so-called \emph{point algebra}.
This language contains relations for $=,<,\leq$, and $\neq$,
interpreted over an infinite dense linear order in the usual way.
Vilain, Kautz, and van Beek showed that consistency of
a given set of constraints over this language (aka the \emph{constraint satisfaction problem} for this language) can be decided 
in polynomial time by local consistency techniques~\cite{PointAlgebra}. 
Later, van Beek described an algorithm that runs in $O(n^2)$, where $n$
is the number of variables~\cite{vanBeek}. 

A considerably larger tractable \tcl\ was 
introduced by B\"urkert and Nebel~\cite{Nebel}. Their language,
called \emph{Ord-Horn}, strictly contains the point algebra.
B\"urkert and Nebel used resolution to show that consistency of a set of 
Ord-Horn constraints can be decided in $O(s^3)$, where $s$ is the
size of the input.
They also showed that establishing path-consistency can
be used to decide whether a given set of Ord-Horn constraints has a solution.
Koubarakis~\cite{Koubarakis} later presented an algorithm
with a running time in $O(s^2)$. 

Ord-Horn is motivated by temporal reasoning tasks for constraints on 
\emph{time intervals}. The study of constraints on intervals (which can be used to model temporal information about events) 
was initiated by Allen~\cite{Allen}, who introduced
an algebra of binary constraint relations on intervals.
The complexity to decide the consistency of a given set of constraints 
from Allen's algebra is NP-complete in general~\cite{Allen}. 
However, for several fragments of
Allen's interval algebra the consistency problem is decidable in
polynomial time. All such fragments
have been classified~\cite{DrakengrenJonsson,KrokhinAllen}.
It is well-known that every relation on \emph{intervals} from Allen's
algebra can be translated to a relation on \emph{time points}.
Hence, algorithmic results for temporal reasoning with time points can be
used for reasoning with time intervals as well.
B\"urkert and Nebel used this
translation to identify one of the tractable fragments of Allen's 
interval algebra, namely the set of all interval constraints 
that translate to Ord-Horn constraints on points.

Another important temporal constraint language with applications in scheduling are AND/OR-precedence constraints~\cite{and-or-scheduling}. 
An AND-constraint can be used to express that some job cannot be started before a set of other jobs has been completed. 
An OR-constraint can be used
to express that a job cannot be started before \emph{one} of
a given set of jobs has been completed.
Feasibility of AND/OR precedence constraints can indeed be
modeled as a constraint satisfaction problem for a \tcl:
AND-constraints 
can be represented by conjunctions of formulas of the form $x > y$,
and OR-constraints by formulas of the form $x > x_1 \vee \dots \vee x > x_n$.


There are temporal constraint languages 
where one cannot expect a polynomial time algorithm.
A well-known \tcl\ with an NP-complete consistency
problem consists of a single ternary relation,  
the \emph{betweenness relation} $\{(x,y,z) \; | \; x{<}y{<}z 
\vee z{<}y{<}x \}$; 
another example of such an NP-complete language consists of
 the \emph{cyclic ordering relation}, which is the ternary relation 
$\{(x,y,z) \; | \; x{<}y{<}z \, \vee \, y{<}z{<}x \, \vee \, z{<}x{<}y\}$.
The constraint satisfaction problems for these two languages
are listed as NP-complete 
in the book of Garey and Johnson~\cite{GareyJohnson}.
We want to remark that the complexity of temporal constraint satisfaction
problems for a \emph{fixed and finite} number of time points 
was completed recently~\cite{finlin}; however, 
the restriction to a finite number
of time points changes the nature of the problem considerably.




We present a new tractable \tcl\ that strictly contains all Ord-Horn
relations and all AND-OR precedence constraints (and also contains relations that are neither Ord-Horn nor AND-OR precedence constraints). Our language
is defined by a universal-algebraic closure property, and 
we call it the class of \emph{$ll$-closed} relations.
We show in Section~\ref{sect:algebraic} that a relation is $ll$-closed 
if and only if it can be defined by a formula of the form
\begin{align*}
(x_1 = y_1 \wedge \dots \wedge x_k = y_k) & \rightarrow (z_0 > z_1 \vee \dots \vee z_0> z_l)
\; \hbox{, or}\\
(x_1 = y_1 \wedge \dots \wedge x_k = y_k) & \rightarrow (z_0 > z_1 \vee \dots \vee z_0> z_l \vee (z_0 = z_1 = \dots = z_l)) 
\end{align*}
(where $k$ and $l$ might be $0$).
It has been shown in~\cite{tcsps} that \X-closed constraints are a \emph{largest tractable language} in the sense
that every
\tcl\ that strictly contains one of our two languages
has an NP-complete constraint satisfaction problem.
The presented algorithm for \X-closed
constraints has a running time that is quadratic in the size of its input.

Traditionally, one of the main algorithmic tools in constraint satisfaction,
and in particular in temporal reasoning, are \emph{local consistency
techniques}~\cite{Allen,PointAlgebra,GolumbicShamir,DrakengrenJonsson,Nebel},
for instance algorithms based on establishing path-consistency. Consistency
based algorithms can be formulated conveniently as Datalog
programs~\cite{BodDal,FederVardi,KolaitisVardi}.  Roughly speaking, Datalog is
Prolog without function symbols, and comes from Database
theory~\cite{EbbinghausFlum}.  We show that, unlike Ord-Horn~\cite{Nebel},
\X-closed and dual \X-closed constraints can \emph{not} be solved by a Datalog
program.  In our proof we apply a pebble-game argument that was originally
introduced for finite domains~\cite{FederVardi,KolaitisVardi}, but has been
shown to generalize to a wide range of infinite domain constraint languages,
including \tcls~\cite{BodDal}.  This is interesting from a
theoretical point of view: for constraint satisfaction problems of languages
over a finite domain, all known algorithms are essentially based on
algebraic algorithms or Datalog~\cite{FederVardi}.  However, the
algorithm we present for temporal reasoning is neither algebraic nor
based on Datalog.

%% file: tcsps.tex
\section{Temporal Constraint Languages}
\label{sect:tcsps}
A \emph{(qualitative) temporal relation} is a relation that is first-order
definable in an unbounded countable dense strict linear order.  All such linear
orders are isomorphic~\cite{Hodges,Marker}, but for convenience we always use
$(\mathbb Q,<)$, i.e., the dense linear order on the rational
numbers\footnote{One could also consider dense linear orders on arbitrary
infinite base sets, e.g.~$(\mathbb R,<)$; but it is easy to see 
all the results in this paper also
apply to this case.}.
An example of a temporal relation
is the ternary \emph{Betweenness} relation $\{(x,y,z) \in {\mathbb Q}^3 \; | \;
(x{<}y \wedge y{<}z) \, \vee \, (z{<}y \wedge y{<}x) \}$ mentioned in the
introduction.  It is well-known that every temporal relation also has a
\emph{quantifier-free} definition~\cite{Hodges,Marker}, i.e., we can define
every temporal relation with a formula that is a Boolean combination of
literals of the form $x<y$ (as above in the case of the Betweenness
relation).


A \emph{temporal constraint language (\tcl)} is an (at most countable) set of
relation symbols $R_1,R_2,\dots$, where each relation symbol $R_i$ is
associated with an arity $k_i \geq 2$, and is interpreted by a $k_i$-ary
temporal relation.  For simplicity, we use the same symbol for the relation
symbol and the corresponding temporal relation.  As an example, consider the
set of binary relation symbols $\Gamma_0 := \{\neq,\leq,<,=\}$, with the obvious interpretation over
$(\mathbb Q,<)$.

The \emph{constraint satisfaction problem} of a temporal
constraint language $\Gamma$ is the following computational
problem.

\cproblem{CSP$(\Gamma)$}
{A first-order formula $\Phi$ of the form
$\phi_1 \wedge \dots \wedge \phi_p \;$, 
where each $\phi_i$ is an atomic 
formula with variables from $x_1,\dots,x_n$ and
a relation symbol from $\Gamma$.}
{Is there an assignment of rational numbers to $x_1, \dots, x_n$ such
that $\Phi$ is satisfied?}

The atomic formulas $\phi_1, \dots, \phi_p$ are called the
\emph{constraints} of the instance $\Phi$ of CSP$(\Gamma)$. 
For a constraint
$\phi = R(x_1,\dots,x_k)$ we say that $\phi$ has \emph{arity}
$ar(\phi) = k$ and for $x_i$ from $\{x_1,\dots,x_k\}$ we say that
$\phi$ is \emph{imposed} on $x_i$.
A tuple $(a_1,\dots,a_n) \in {\mathbb Q}^n$ is called a \emph{solution} for
$\Phi$ if the assignment $x_i:=a_i$ satisfies all formulas in $\Phi$.
If there is no solution for $\Phi$, then we say that $\Phi$ is 
\emph{unsatisfiable}, and \emph{satisfiable} 
(or \emph{consistent})  otherwise. Thus, 
CSP$(\Gamma)$ is the problem to decide whether a given
set of constraints over relations from $\Gamma$ is satisfiable. \\

\noindent{\em Example.} Let $R(x,y,u,v)$ be the $4$-ary temporal relation defined
by $(x{=}y \wedge y{<}u \wedge u{=}v) \vee (x{<}y \wedge y{<}u \wedge
u{<}v)$. Consider the instance $\Phi_1 := \{R(x_1,x_2,y_1,y_2),
R(x_1,x_2,y_2,y_3),R(x_1,$ $x_2,y_3,y_1)\}$ of CSP$(\{R\})$. It is
easy to see that the sentence $\exists x_1,x_2,$ $y_1,y_2,y_3
\bigwedge_{\phi \in \Phi_1} \phi$ is true, and a solution to $\Phi_1$
is $(0,0,1,1,1)$. \\

A \emph{finite} constraint language $\Gamma$ is called \emph{tractable} if CSP($\Gamma$) can be solved in polynomial time (note that because $\Gamma$ is finite this concept is independent from 
the representation of the relation symbols in the input). The
constraint language $\Gamma_0$ mentioned at the beginning of this section, for example, corresponds to the well-studied point-algebra that we mentioned in the introduction, and is tractable. 
An \emph{infinite} constraint
language $\Gamma$ is called \emph{locally tractable} if every finite subset of the constraint language is tractable. 
The algorithmic results presented in this paper show more than
local tractability of constraint languages.
To formulate our results, we have to discuss how to represent
temporal relations in instances of CSP$(\Gamma)$
when $\Gamma$ is infinite.

It is straightforward to verify that whether or not an $n$-tuple $t$ is in
a temporal relation only depends on the weak linear order $tp(t)$ defined on $\{1,\dots,n\}$ by $(i,j) \in tp(t)$ iff $t[i] \leq t[j]$. We also say that $t$ \emph{satisfies} $tp(t)$~\footnote{The notation $tp$ is motivated by the
concept of \emph{(complete) types} in model theory~\cite{Hodges,Marker}.}.
This observation leads to a natural way to represent temporal relations.  If
$R$ is a $k$-ary temporal relation, $R$ can be represented by a set $\cal R$ of
weak linear orders on $\{1,\dots,k\}$ as follows.  For every $k$-tuple $t \in
R$, the weak linear order $tp(t)$ is contained in $\cal R$. Conversely, for
every weak linear order $w$ in $\cal R$ there is a $k$-tuple $t \in R$ such
that $w=tp(t)$. For example, the relation $R$ in the example above can be
characterized as the set of all tuples that satisfy either $tp((0,0,1,1))$ or
$tp((0,1,2,3))$.

If $\Gamma$ is the set of all temporal relations, 
then CSP$(\Gamma)$ is well-known to be NP-complete (here we assume that  temporal relations are represented by sets of weak linear orders). For containment in NP, note that one can verify
in polynomial time whether a given weak linear order on $n$ variables
corresponds to a solution for a given instance $\Phi$ with $n$ variables.  We
can therefore decide non-deterministically in polynomial time whether there
exists a weak linear order on $n$ elements and an (arbitrary) $n$-tuple $t$
satisfying this weak linear order such that $t$ is a solution to $\Phi$.  For
NP-hardness, recall that already the constraint language that only contains a
single relation symbol for the Betweenness relation is
NP-complete~\cite{GareyJohnson}.

For a fixed way how to represent the relations from a constraint
language $\Gamma$ (such as the representation of temporal relations by sets of weak linear orders as discussed above), we say that $\Gamma$
is \emph{(globally) tractable} if CSP$(\Gamma)$ can be solved
in polynomial time. 
The representation of a temporal relation by a set of weak linear orders corresponds to the standard representation of a relation over a finite domain by its set of tuples\footnote{Also for finite domains, a representation by formulas might sometimes be more natural, for example for boolean Horn satisfiability.}. 
For finite domains, it is an open problem whether the notion of local tractability and the notion of global tractability with respect to the standard representation coincide (and in fact it has been conjectured that they do~\cite{JBK}).

Another natural way to represent a temporal 
relation is by specifying a formula
that defines the relation. However, general Boolean combinations of literals of the form $x<y$ are obviously too expressive if we are interested
in efficient algorithms (it is NP-hard to decide whether such a
formula represents a non-empty relation), so we
have to restrict the set of all formulas appropriately;
such a syntactic restriction will be presented in Section~\ref{sect:algebraic}.

In this paper we present an algorithm that shows that a large temporal constraint language is globally tractable, both with respect to the representation of constraint relations by sets of weak linear orders
and with respect to the representation of temporal relations by formulas.
Even though this requires that we have to go into more detail as compared to a local tractability result, we would like to present the stronger global tractability result in this paper, because
this allows us to relate our algorithm with previous algorithms in temporal reasoning, for example for Ord-Horn (which is an
infinite temporal constraint language).



%% file: algebraic.tex
\section{\X-closed Languages}\label{sect:algebraic}
We first introduce fundamental concepts from model theory and universal
algebra; they are standard, see e.g.~\cite{Hodges,Szendrei}. 
We say that a $k$-ary function (also called \emph{operation}) 
$f: {\mathbb Q}^k \rightarrow {\mathbb Q}$ \emph{preserves} an $m$-ary relation 
$R \subseteq {\mathbb Q}^m$ if whenever $R(a_1^i, \dots, a^i_m)$ 
holds for all $1 \leq i \leq k$, 
then $R\big(f(a_1^1, \dots, a_1^k), \dots, f(a_m^1, \dots,
a_m^k) \big)$ holds as well.
If $f$ preserves all relations of a \tcl\ $\Gamma$,
we say that $f$ is a \emph{polymorphism} of $\Gamma$.
Unary bijective polymorphisms are called the \emph{automorphisms} 
of $\Gamma$; the set of all automorphisms of $\Gamma$ is denoted by Aut$(\Gamma)$.


Let \lex\ be a binary operation on ${\mathbb Q}$ such that $\lex(a,b) <
\lex(a',b')$ if either $a < a'$, or $a=a'$ and $b<b'$.  It is easy to see that
the set of temporal relations preserved by \lex\ is not affected by the choice
of the binary operation \lex\ if \lex\ has the properties above. Thus for all
the arguments in this paper, it does not matter which operation \lex\ we are
chosing.  Also note that every such operation is by definition injective.

\begin{figure}
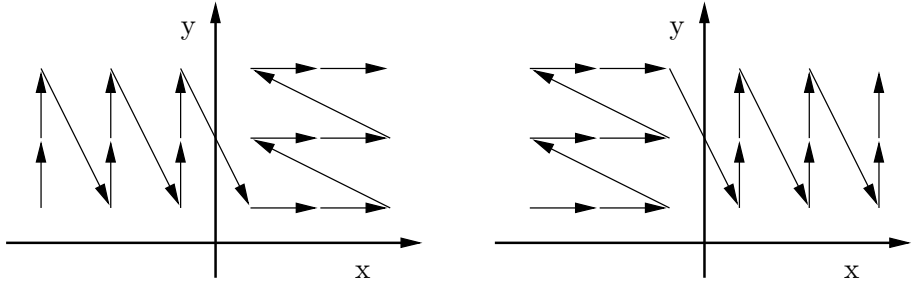

\begin{center}
\input ll_dualll.pstex_t
\end{center}
\caption{A visualization of the ll (left side) and the dual ll operation (right side).}
\label{fig:ll_dualll}
\end{figure}

Let $\lele$ be a binary operation on ${\mathbb Q}$ 
such that $\lele(a,b) < \lele(a',b')$ if one of the following cases applies.
\begin{itemize}
\item $a \leq 0$ and $a<a'$
\item $a \leq 0$ and $a=a'$ and $b<b'$
\item $a,a' > 0$ and $b<b'$
\item $a > 0$ and $b=b'$ and $a<a'$
\end{itemize}
See Figure~\ref{fig:ll_dualll} for illustration. In diagrams like in Figure~\ref{fig:ll_dualll} we draw a directed edge from $(a,b)$ to $(a',b')$ 
if $\lele(a,b) < \lele(a',b')$. 
Again, it is easy to see that
the set of temporal relations preserved by \lele\ is not affected by the 
exact choice of the binary operation \lele.
Also observe that every temporal relation that is preserved by 
\lele\ is also preserved by \lex. 
We say that a relation is \X-closed if it
is preserved by \lele.

It is possible to decide algorithmically whether a constraint language is
\lele-closed. 

\begin{proposition}\label{prop:meta}
Given a finite constraint language where all relations are
represented as lists of weak linear orders, one can decide in
polynomial time in the input size whether the constraint language
is \lele-closed.
\end{proposition}

\begin{proof}
We test for each relation $R$ in the constraint language separately whether it
is \X-closed.  A $k$-ary relation $R$ is preserved by \lele\ if and only if for
every two weak orders $o_1$ and $o_2$ in $R$ and every index $e\leq k$ the weak
order $o_3$ is also in $R$, where $o_3$ is defined as follows: $(i,j) \in o_3$
iff one of the following holds
\begin{itemize}
\item $(i,j) \in o_1$ and $(i,j) \in o_2$, 
\item $(i,j) \in o_1$, $(j,i) \notin o_1$, and $(i,e) \in o_1$, or
\item $(i,j) \in o_2$, $(j,i) \notin o_2$, $(e,j) \in o_1$, and $(j,e) \notin o_1$.
\end{itemize}
For all pairs $(o_1,o_2)$ of weak linear orders on $\{1,\dots,k\}$
in the representation of $R$, and for each index $e \leq k$,
we can verify in linear time in $k$ whether the weak linear order $o_3$ 
as described above is also contained in the representation of $R$.
\end{proof}

Similarly to the \lele\ operation we can define a dual \lele\ operation, as
depicted in Figure~\ref{fig:ll_dualll}.  To show that the language of all \X-closed relations is different from the language of all dual \lele-closed relations, we use the following two relations, which will be also of importance in later arguments.

\begin{definition}
We define $R^{min}$ to be the ternary relation 
$\{ (x,y,z) \; | \; x {>} y \vee x {>} z \}$, 
and $R^{max}$ to be $\{ (x,y,z) \; | \; x {<} y \vee x {<} z\}$.
\end{definition}
Observe that $R^{min}(x,y,z)$ holds if and only if $x$ is larger than 
the minimum of $y$ and $z$. Similarly, $R^{max}(x,y,z)$ holds if and
only if $x$ is smaller than the maximum of $y$ and $z$.
It was shown in~\cite{and-or-scheduling} and independently in~\cite{GM2} that CSP$(\mathbb Q, R^{max})$
can be solved in polynomial time. 
For the proof of the next lemma we prove that the relation $R^{min}$
is $ll$-closed; the proof can be adapted easily to show that all $k$-ary relations defined by formulas of the form $x_1 > x_2 \vee \dots \vee x > x_k$  are $ll$-closed  as well, which are the relations to model AND/OR precedence constraints.

\begin{proposition}\label{prop:incomp}
The language of \X-closed relations does not contain the class of dual
\lele-closed relations and vice versa.
\end{proposition}
\begin{proof}
To show that the language of \X-closed constraints does not contain the language of dual \lele-closed constraints,
we show that there is a temporal relation that is preserved
by \lele\ but not by dual \lele.
We claim that the relation $R^{min}$ is preserved by the \lele\ operation:
Let $(x_1,x_2,x_3)$ and $(y_1,y_2,y_3)$ be triples that are both in
the relation $R^{min}$. 
Without loss of generality, $x_1 > x_2$ 
(note that the relation is symmetric in the second and third argument).
If in this case $y_1 \geq y_2$, then, because ll preserves $\leq$,
we have that $\text{ll}(x_1,y_1) \geq \text{ll}(x_2,y_2)$, 
and because ll is injective,
we have that $\text{ll}(x_1,y_1) > \text{ll}(x_2,y_2)$. Therefore 
$(\text{ll}(x_1,y_1),\text{ll}(x_2,y_2),\text{ll}(x_3,y_3))$ 
is in $R^{min}$, and we are done.
So let us assume that
$y_1 < y_2$ and therefore $y_1 > y_3$.
We can again apply
the previous argument to show that $(\text{ll}(x_1,y_1),\text{ll}(x_2,y_2),\text{ll}(x_3,y_3))$ is in $R^{min}$
unless $x_1 < x_3$. So let us assume that $x_1 < x_3$.
Now, in case that $x_2 > 0$, the operation ll 
preserves $R^{min}$, since in this case ll acts like a lexicographic 
order on the two triples. 
Otherwise, $x_2 \leq 0$. 
It is easy to check that then $\text{ll}(x_2,y_2)<\text{ll}(x_1,y_1)$
because $x_1 > x_2$.

However, $R^{min}$ is not preserved by the dual ll operation:
consider the tuples $t_1:=(-1,1,-2)$ and $t_2:=(-1,-2,1)$ 
that are both in $R^{min}$. If we apply the dual ll operation 
to these two tuples, 
we obtain $\text{dual-ll}(-1,-1) < \text{dual-ll}(-2,1) < \text{dual-ll}(1,-2)$, and hence the
tuple $\text{dual-ll}(t_1,t_2)$ is not in the relation $R^{min}$. 

This shows that the language of \X-closed constraints does not contain the
language of dual \lele-closed constraints.  Analogously, we can use the
relation $R^{max}$ to show that the language of dual \lele-closed constraints
does not contain the language of \lele-closed constraints.
\end{proof}

The temporal constraint language of all \lele-closed relations also contains the
important class of Ord-Horn relations, introduced by B\"urckert and
Nebel~\cite{Nebel} to identify a tractable class of interval constraints.
A relation is Ord-Horn if it can be defined by a conjunction of formulas
of the form
\begin{align*}
(x_1=y_1 \wedge \dots \wedge x_k=y_k) \rightarrow x_0 \; O \; y_0 \; ,
\end{align*}
where $O \in \{=,<,\leq,\neq\}$. It is always possible to translate interval
constraints into temporal constraints~\cite{PointAlgebra}.  If the translation
of an interval constraint language falls into a tractable \tcl, the interval
constraint language is tractable as well.  B\"urckert and Nebel showed that the
class of \emph{interval} constraints having a translation into Ord-Horn
temporal constraints is a largest tractable fragment of Allen's interval
algebra.  Note that this does not imply that the class of Ord-Horn constraints
is a largest tractable \tcl\ on time points.  Indeed, this is not the case.
Proposition~\ref{prop:contains} below shows that the class of Ord-Horn
constraints is \X-closed. Since the relation $R^{min}$ defined in this section
is \X-closed but not Ord-Horn, the class of \X-closed constraints is
\emph{strictly} larger than Ord-Horn.  Finally, we prove in
Section~\ref{sect:alg} that \X-closed constraints are tractable.


\begin{proposition}\label{prop:contains}
All relations in Ord-Horn are preserved by ll and by dual ll.
\end{proposition}
\begin{proof}
We will give the argument for the ll operation only; the argument for
the dual ll operation is analogous.
It suffices to show that every relation 
that can be defined by a formula $\Phi$ 
of the form $(x_1=y_1 \wedge \dots \wedge x_{k-1}=y_{k-1}) \rightarrow x_k \; O \; y_k$ is preserved by ll, where $O \in \{=,<,\leq,\neq\}$.
Let $t_1$ and $t_2$ be two $2k$-tuples that satisfy
$\Phi$. Consider a $2k$-tuple $k_3$ obtained by applying
ll componentwise to $t_1$ and $t_2$. We distinguish two cases:
either there is an $i \leq k-1$ such that in one of the tuples 
$x_i = y_i$ is not satisfied -- in this case $x_i = y_i$ is not satisfied
in $t_3$ as well by injectivity of ll, 
and therefore the tuple $t_3$ satisfies $\Phi$.
Or $x_i=y_i$ holds for all $i\leq k-1$ in both tuples $t_1$ and $t_2$. 
But then, as $t_1$ and $t_2$ satisfy $\Phi$, the literal $x_k O y_k$ holds 
in both $t_1$ and $t_2$. Since ll preserves all relations in $\{=,<,\leq,\neq\}$, 
the literal $x_k O y_k$ holds in $t_3$, and
therefore $t_3$ satisfies $\Phi$ as well.
\end{proof}


It turns out that a temporal relation is preserved by ll if and only if
it can be defined by a class of formulas which we call \emph{ll-Horn formulas}.
This class properly extends the class of Ord-Horn formulas. 
A formula is called \emph{ll-Horn} if it is a conjunction of formulas
of the following form (slightly abusing terminology, we call 
these formulas the \emph{clauses} of the ll-Horn formula)
\begin{align*}
(x_1 = y_1 \wedge \dots \wedge x_k = y_k) & \rightarrow (z_0 > z_1 \vee \dots \vee z_0> z_l)
\; \hbox{, or}\\
(x_1 = y_1 \wedge \dots \wedge x_k = y_k) & \rightarrow (z_0 > z_1 \vee \dots \vee z_0> z_l \vee (z_0 = z_1 = \dots = z_l)) 
\end{align*}
where $0 \leq k,l$. Note that $k$ or $l$ might be $0$: if $k=0$, we obtain
a formula of the form $z_0 > z_1 \vee \dots \vee z_0> z_l$ or $(z_0 > z_1 \vee \dots \vee z_0> z_l \vee (z_0 = z_1 = \dots = z_l))$,
and if $l=0$ we obtain a disjunction of disequalities.
Also note
that the variables $x_1,\dots,x_k,y_1,\dots,y_k,z_0,$
$\dots,z_l$ need not be pairwise distinct. 
Also note that the clause $z_1 > z_2 \vee z_3 > z_4$ is \emph{not} equivalent to an ll-Horn formula.

\begin{proposition}\label{prop:ll-horn}
A temporal relation is ll-closed if and only if it can be defined by an ll-Horn formula.
\end{proposition}

We first prove Lemma~\ref{lem:inj} below about relations that only contain \emph{injective} tuples.
A tuple is said to be \emph{injective} if all entries of the tuple
are pairwise distinct. 
Note that every temporal relation $R$
can be defined by a quantifier-free formula in conjunctive normal
form where all literals are of the form $x > y$ or $x=y$;
to see this, take any quantifier-free formula in conjunctive
normal form that defines $R$ and 
\begin{itemize}
\item replace $x \leq y$ by the two literals $x<y \vee x=y$;
\item replace $x \neq y$ by the two literals $x > y \vee y > x$;
\item replace $x < y$ by $y > x$.
\end{itemize}
We call formulas in quantifier-free conjunctive normal form where all literals are of the form $x>y$ or $x=y$ \emph{standard formulas}.
A clause is \emph{bad} if it is not of the form $z_0 > z_1 \vee \dots \vee z_0> z_l$.

\begin{lemma}\label{lem:inj}
Let $R$ be a temporal relation that only contains injective tuples, and let $\phi$ be a standard formula with minimal number of bad clauses such that
\begin{itemize}
\item[a)] an injective tuple is in $R$ if and only if it satisfies $\phi$;
\item[b)] any formula obtained from $\phi$ by removing a literal from a clause
does not have this property.
\end{itemize}
If $R$ is $ll$-closed, then $\phi$ does not contain bad clauses.
\end{lemma}

\begin{proof}
Suppose that $R$ is $n$-ary, and that $x_1,\dots,x_n$ are the variables of $\phi$. Because $R$ only contains injective tuples, 
we can remove literals of the form $y=z$ from $\phi$. But this would contradict
assumption a),
so we assume that $\phi$ only contains literals of the form $x > y$.

Suppose for contradiction that $\phi$ contains a bad clause $C$.
Then $C$ must contain two literals 
$l_1 := x_u > x_v$ and $l_2 := x_r > x_s$ where $x_u$ and $x_r$ are distinct variables. 
We claim that there is an injective 
tuple $t_1$ such that $l_1$ is the only literal satisfied in $C$ if we assign $t_1[i]$ to $x_i$ for $1 \leq i \leq n$. Otherwise, the formula obtained
from $\phi$ by removing $l_1$ from $C$ still has the property
that every injective tuple is in $R$ if and only if it satisfies $\phi$.
Moreover, the number of bad clauses in the new formula is also minimal,
which is impossible by the choice of $\phi$.
Similarly one can see that there is an injective tuple $t_2$ such that $l_2$ is the only literal satisfied in $C$ if we assign $t_2[i]$ to $x_i$.

We first study the case that 
$t_1$ can be chosen such that $t_1[r]$ is smaller than $t_1[s]$, $t_1[u]$, and $t_1[v]$. Let $\alpha$ be an automorphism of
$({\mathbb Q},<)$ such that $t_1[r]$ is mapped to $0$. Consider
the tuple $t = ll(\alpha(t_1),t_2)$. Observe that $t$ is injective
since \X preserves $\neq$. If a literal
in some clause of $\phi$ is not satisfied in both tuples $t_1$ and $t_2$,
then it is also not satisfied in $t$, because \X and $\alpha$ preserve
$\leq$. 
Therefore only the literals $l_1$ and $l_2$ of $C$ can be satisfied by
$t$. Since $t[r]$ is strictly smaller than $t[s]$ (by the properties of \X), 
the literal $l_2$ cannot be satisfied by $t$ in $C$. Since $t_2[v] > t_2[u]$, it also holds
that $t[v]>t[u]$ (by the properties of \X), 
and hence $l_1$ is not satisfied in $t$ either.
So $t$ does not give a satisfying assignment for $\phi$, 
in contradiction with the assumption that $R$ is \X-closed.

An analogous argument shows that $t_2$ cannot be chosen such 
that $t_2[u]$ is smaller than $t_2[v]$, $t_2[r]$, and $t_2[s]$. 
We claim that any injective tuple that satisfies $\phi$ also satisfies
 $(x_u > x_v \vee x_u > x_s)$.
If there was an injective tuple $t$ with $t[u] < t[v]$ and $t[u] < t[s]$,
then $t[u]>t[r]$ to satisfy the property from the beginning of the paragraph. Hence, $t[r]<t[v]$ and $t[r] < t[s]$.
But then $t[r]$ is smaller than $t_2[v]$, $t_2[r]$, and $t_2[s]$,
in contradiction to what we have shown before.
Analogously we can show that any injective tuple that satisfies
$\phi$ also satisfies $(x_r > x_v \vee x_r > x_s)$.

Let $\phi'$ be the formula obtained from $\phi$ by removing $C$ and adding
these two clauses. We show that an injective tuple satisfies $\phi'$ 
if and only if it satisfies $\phi$. By what we have see above, 
it suffices to show that $\phi'$ implies $\phi$.
Let $t$ be any satisfying assignment of $\phi'$. Clearly,
all the clauses of $\phi$ except for $C$ are satisfied by $t$, because
they are also present in $\phi'$. 
We can reformulate the two additional clauses in $\phi'$ to 
$$ (x_u > x_v \wedge x_r > x_v) \vee (x_u > x_v \wedge x_r > x_s)
\vee (x_u  > x_s \wedge x_r > x_v) \vee (x_u > x_s \wedge x_r > x_s) \; .$$ 
If the first, the second, or the fourth disjunct is satisfied by $t$, then 
$t[x_u] > t[x_v] \; \vee \; t[x_r] > t[x_s]$, and therefore $C$ holds in $t$.
If the third disjunct is satisfied by $t$ and the literal $l_1$ does not
hold (i.e., $t[x_u]<t[x_v]$), we have the chain of inequalities $t[x_s]<t[x_u] < t[x_v] < t[x_r]$ and hence $t[x_r] > t[x_s]$. 
Thus, also in this last case $C$ holds.

The formula $\phi'$ has fewer bad clauses than $\phi$.
Let $\phi''$ be the formula obtained from $\phi'$ by 
repeatedly removing literals from clauses as long as 
an injective tuple is in $R$ if and only if it satisfies $\phi''$.
Since removing literals does not create new bad clauses, we eventually
obtain a formula that contradicts the choice
of $\phi$. 

We thus have shown that $\phi$ cannot contain bad clauses.
\end{proof}

\begin{proof}[Proof of Proposition~\ref{prop:ll-horn}]
The proof that every relation defined by an ll-Horn formula is ll-closed
is similar to the proof of Proposition~\ref{prop:contains}.
We just need to additionally check that the relation 
defined by $z_0 > z_1 \vee \dots \vee z_0 > z_l$
and the relation defined by $z_0>z_1 \vee \dots \vee z_0 > z_l \vee 
(z_0 = \dots = z_l)$ are preserved by ll, which is straightforward.

The proof of the reverse implication is by induction on the arity
$n$ of the temporal relation $R$. We assume that $R$ is ll-closed.
For $n=2$ the statement of the proposition holds, because all binary temporal
relations can be defined by ll-Horn formulas.
For $n>2$, we construct the formula $\psi$ that defines $R$ as follows.


Let $\phi$ be a standard formula with minimal number of bad clauses such that
a) an injective tuple is in $R$ if and only if it satisfies $\phi$, and b) 
any formula obtained from $\phi$ by removing a literal from a clause
does not satisfy condition a). 
Clearly, such a formula exists: we can start from
any standard formula that defines $R$ and has a minimal number
of bad clauses, and then remove repeatedly literals from clauses 
if the resulting formula still satisfies a); since deleting literals
does not create bad clauses, we eventually find a formula that satisfies
both conditions a) and b). 
Lemma~\ref{lem:inj} shows that $\phi$ does not contain
bad clauses.

For all pairs of entries $i,j \in \{1,\dots,n\}$, $i<j$, 
let $R_{i,j}$ be the projection
of the relation $R(x_1,\dots,x_{i-1},x_j,x_{i+1},\dots,x_n)$ 
to $x_1,\dots,x_{j-1},x_{j+1},\dots,x_n$. Because also $R_{i,j}$ is ll-closed, it has an ll-Horn definition
$\phi_{i,j}$ by inductive assumption.
We add to each clause of $\phi_{i,j}$ a literal $x_i=x_j$
to the premise of the implication, such that $\phi_{i,j}$ remains
an ll-Horn formula. 

Let $\psi$ be the formula that is a conjunction of
\begin{itemize}
\item all the modified clauses from all formulas $\phi_{i,j}$;
\item all clauses $C(z_0,\dots,z_l)$ of $\phi$ 
such that $R$ does not contain a tuple where $z_0,z_1,\dots,z_l$ all
get the same value;
\item the formula $C(z_0,\dots,z_l) \vee (z_0=z_1=\dots=z_l)$ for all other clauses $C$ of $\phi$ with variables $z_0,z_1,\dots,z_l$.
\end{itemize}
Obviously, $\psi$ is an ll-Horn formula. We have to verify that
$\psi$ defines $R$. Let $t$ be an $n$-tuple such that $t \notin R$.
If $t$ is injective,  then some 
clause $C(z_0,z_1,\dots,z_l)$ of $\phi$ is not satisfied by $t$. 
The variables $z_0,z_1,\dots,z_l$ of $C$ cannot all have the
same value in $t$, and so $\psi$ is not satisfied either. 
If there are $i,j$ such that $t[i]=t[j]$ then the tuple $t_j=(t[1],\dots,t[j-1],t[j+1],\dots,t[n]) \notin R_{i,j}$. Therefore some clause $C$ of 
$\phi_{i,j}$ is not satisfied by $t_j$, and $C \vee x_i \neq x_j$
is not satisfied by $t$. Thus, in this case $t$ does not satisfy $\psi$, too.

We also have to verify that all $t \in R$ satisfy $\psi$.
Let $C$ be a conjunct of $\psi$ created from some clause in $\phi_{i,j}$. If $t[i] \neq t[j]$, then $C$ is satisfied by $t$ because $C$
contains $x_i \neq x_j$. If $t[i] = t[j]$, then $(t[1],\dots,t[j-1],t[j+1],\dots,t[n]) \in R_{i,j}$ and thus this tuple satisfies $\phi_{i,j}$. This also implies that $t$ satisfies $C$. 

Finally, let $C$ be a conjunct of $\psi$ created from some clause of $\phi$. Then $C$ is of the form
$x_{u_0} > x_{u_1} \vee \dots \vee x_{u_0} > x_{u_m}$ 
or of the form 
$x_{u_0} > x_{u_1} \vee \dots \vee x_{u_0} > x_{u_m} \vee (x_{u_0}=x_{u_1}=\dots=x_{u_m})$. 
If $t$ is constant on the variables of $C$, then,
by construction of $\psi$, $C$ contains the disjunct 
$x_{u_0}=x_{u_1}=\dots=x_{u_m}$ and
is satisfied. So suppose $t$ is not constant on the variables of $C$.
Assume for contradiction 
that $t[u_0] \leq t[u_i]$ for all $i \in [m]$. 
Since $t$ is not constant, there is a
$j \in [m]$ such that $t[u_0]<t[u_j]$. By our assumptions on $\phi$, 
there is an injective tuple $t' \in R$ that
satisfies only the literal $x_{u_0} > x_{u_j}$ in $C$. Consider 
the tuple $t'' = {\it lex}(t,t')$. Because $t'$ has pairwise distinct entries
and lex is an injective operation, $t''$ also has pairwise distinct entries. Since $t[u_0]  \leq t[u_i]$ for all $i \in [m]$, $t'[u_0]<t'[u_i]$
for all $i \in [m] \setminus \{j\}$, and because lex
preserves $\leq$, we get that $t''[u_0] < t''[u_i]$ for
all $i \in  [m] \setminus \{j\}$. Finally, since $t[u_0]<t[u_j]$
we also have that $t''[u_0] < t''[u_j]$ by the properties of lex. 
Hence, $t''[u_0] < t''[u_i]$ for all $i \in [m]$. Therefore
$t''$ does not satisfy $C$ and thus also does not satisfy $\phi$. 
But $t''$ is injective and is from $R$ (because $R$ is ll-closed),
in contradiction to the properties of $\phi$.
\end{proof}

The syntactic characterization of ll-closed temporal relations 
motivates our the second way of representing the constraints
in the input instances of the constraint satisfaction problem for
ll-closed temporal constraint languages --- the constraints may be given as ll-Horn formulas. 
We would like to remark that there are ll-closed temporal relations where
the tuple representation is more succinct than the
ll-Horn representation, and vice versa.
Note that since every Ord-Horn formula is obviously 
an ll-Horn formula, the algorithm we present for
the ll-Horn representation in Section~\ref{sect:alg} strictly
generalizes the existing algorithms for Ord-Horn constraints.
For simplicity, we assume that in instances of the constraint satisfaction
problem for ll-closed \tcls\ 
the formulas representing the constraints consist of just one clause 
(we can always transform a
constraint into several constraints of this form).

\ignore{
\begin{proposition}
Temporal relations that can be defined by an ll-Horn formula are
\X-closed.
\end{proposition}

In fact, it can be shown that the converse is true as well:
if a temporal relation is ll-closed it can be defined by an ll-Horn formula; the proof of this fact (which we do not need here)
will appear in the long version of this paper.
}

%% file: ll_dualll.pstex_t
\begin{picture}(0,0)%
\includegraphics{ll_dualll.pstex}%
\end{picture}%
\setlength{\unitlength}{4144sp}%
\begingroup\makeatletter\ifx\SetFigFont\undefined%
\gdef\SetFigFont#1#2#3#4#5{%
  \reset@font\fontsize{#1}{#2pt}%
  \fontfamily{#3}\fontseries{#4}\fontshape{#5}%
  \selectfont}%
\fi\endgroup%
\begin{picture}(5475,1715)(741,-1706)
\put(4732,-222){\makebox(0,0)[lb]{\smash{{\SetFigFont{11}{13.2}{\rmdefault}{\mddefault}{\updefault}{\color[rgb]{0,0,0}y}%
}}}}
\put(5776,-1684){\makebox(0,0)[lb]{\smash{{\SetFigFont{11}{13.2}{\rmdefault}{\mddefault}{\updefault}{\color[rgb]{0,0,0}x}%
}}}}
\put(1807,-222){\makebox(0,0)[lb]{\smash{{\SetFigFont{11}{13.2}{\rmdefault}{\mddefault}{\updefault}{\color[rgb]{0,0,0}y}%
}}}}
\put(2852,-1684){\makebox(0,0)[lb]{\smash{{\SetFigFont{11}{13.2}{\rmdefault}{\mddefault}{\updefault}{\color[rgb]{0,0,0}x}%
}}}}
\end{picture}%

%% file: algorithm.tex
\section{An Algorithm for \X-closed Constraints}
\label{sect:alg}
\newcommand{\KK}{\cal{K}}
\newcommand{\CC}{\cal{C}}


In this section we present an algorithm for ll-closed constraints.  It is
straightforward to `dualize' the algorithm and all arguments, and we will
therefore also obtain an algorithm for dual ll-closed constraints.

One of the underlying ideas of the algorithm is to use a subroutine
that tries to find a solution where every variable
has a different value. If this is impossible, the subroutine must
return a set of at least two variables that denote the same
value in all solutions. It is one of the fundamental properties
of ll-closed constraints that this is always possible.


To formally introduce our algorithm we need the following definitions.  Let
$\phi=R(x_1,\dots,$
$x_k)$ be an atomic formula where $R$ is a temporal relation
that is preserved by an operation $f$.  Clearly, for all $x_i$ from
$x_1,\dots,x_k$ the temporal relation defined by $\exists x_i. \phi$ is
preserved by $f$ as well.  Therefore, if $\Phi$ is an instance of the CSP with
constraints that are preserved by $f$, and $\overline y$ is a sequence of some
of the variables of $\Phi$, then $\Phi' := \{ \exists \overline y. \phi \; | \;
\phi \in \Phi\}$ can also be viewed as an instance of the CSP with constraints
preserved by $f$.  We call $\Phi'$ the \emph{projection} of $\Phi$ to $X
\setminus \overline y$.  Note that if $\Phi'$ is unsatisfiable, then $\Phi$ is
unsatisfiable as well.

The $i$-th entry in a $k$-tuple $t$ is called {\em minimal} 
if $t[i]\leq t[j]$ for every $j\in[k]$. 
It is called {\em strictly minimal} if $t[i]<t[j]$ for every $j\in[k]
\setminus\{i\}$.

\begin{definition}
Let $R$ be a $k$-ary relation. A set $S\subseteq[k]$ is
called a \emph{min-set for the $i$-th entry in $R$} if there exists a tuple
$t\in R$ such that the $i$-th entry is minimal in $t$, 
and for all $j \leq k$ it holds that $j \in S$ if and only if $t[i]=t[j]$. 
We say that $t$ is a \emph{witness} for this min-set.
\end{definition}

Let $R$ be a $k$-ary relation 
that is preserved by \lex\ 
(recall that ll-closed constraints are preserved by \lex\ as well),
and suppose that the $i$-th entry has 
the min-sets $S_1,\ldots,S_l$, for $l \geq 1$, 
with the corresponding witnesses $t_1,\dots,t_l$. Consider the tuple
$t:=\lex(t_1,\lex(t_2,\ldots$ $\lex(t_{l-1},t_l)))$. Since the entry $i$ 
is minimal in every tuple $t_1,\dots,t_l$, and since \lex\ preserves 
both $<$ and $\leq$, it is also minimal in $t$. 
Because \lex\ is injective, we have that $t[i]=t[j]$ 
if and only if these two entries are equal in each tuple $t_1,\dots,t_l$. 
Hence, the min-set for the $i$-th entry in $R$ witnessed by
the tuple $t$ is a subset of every other min-set $S_1,\ldots,S_l$.
We then call this set the {\em minimal min-set} for the $i$-th entry in $R$.

\begin{lemma}
\label{lem:lex-free-set-equal}
Let $R$ be a $k$-ary relation preserved by \lex, $t\in R$, $i\in[k]$
and $S$ be the minimal min-set for the $i$-th entry in $R$.
If $t$ is such that $t[j] \geq t[i]$ for every $j \in S$,
then $t[i]=t[j]$ for every $j\in S$.
\end{lemma}
\begin{proof}
Let $t' \in R$ be the tuple that witnesses the minimal min-set $S$. 
Suppose there is a tuple $t \in R$ such that not all entries in $S$ are equal
(in particular, $|S| > 1$). 
Consider the tuple $t'':=lex(t',t)$. By the properties of \lex\ 
it holds that $t''[i] < t''[j]$ 
for every $j\in[k]\setminus S$. Furthermore, 
$t''[i]\leq t''[j]$
for $j\in S$ if and only if $t[i]\leq t[j]$. Thus, unless $t''$ witnesses
a smaller min-set for $i$ in $R$ (which would be a contradiction), we have 
that $t''[i]>t''[j]$ for some $j\in S$.
\end{proof}

To develop our algorithm, we use a specific notion of \emph{constraint graph} 
of a temporal CSP instance, defined as follows.

\begin{definition}
The \emph{constraint graph} $G_\Phi$ of a temporal CSP instance $\Phi$ is a
directed graph $(X,E)$ defined on the variables $X$ of $\Phi$.
For each constraint of the form $R(x_1,\dots,x_k)$ from $\Phi$ we add
a directed edge $x_i x_j$ to $E$ if in every tuple from $R$ where the $i$-th entry is minimal the $j$-th entry is minimal as well.
\end{definition}

{\bf Example.} We return to the example from Section~\ref{sect:tcsps}.
The constraint graph $G_{\Phi_1}$ for the instance $\Phi_1$ in this example
has the vertices $x_1,x_2,y_1,y_2,y_3$, 
edges from each of $y_1,y_2,y_3$ to all other variables, and an
edge from $x_2$ to $x_1$.


\begin{definition}
If $\Phi$ contains a constraint $\phi$ imposed on $y$
such that $\phi$ does not admit a solution where $y$ denotes the minimal value,
the we say that $y$ is \emph{blocked (by $\phi$)}.
\end{definition}

We can easily determine for each constraint which variables are blocked by this
constraint: For a constraint represented by weak linear orders we just check all
weak linear orders and build a set of variables that are not minimal in any of them.
For a constraint represented by an ll-Horn formula, a variable $x_i$ is blocked if and only if the formula is of the form 
$x_i > z_1 \vee \dots \vee x_i > z_l$.
Thus, by
inspecting all the constraints it is possible to compute the blocked variables
in linear time in the input size. We would like to use the constraint graph to
identify variables that have to denote the same value in all solutions, and
therefore introduce the following concepts.

\begin{definition} 
A strongly connected component $K$ of the constraint graph $G_\Phi$ for a
temporal CSP instance $\Phi$ is called a \emph{sink component} if no edge in
$G_\Phi$ leaves $K$, and no variable in $K$ is blocked.  A vertex of $G$ that
belongs to a sink component of size one is called a \emph{sink}.
\end{definition}

{\bf Example.} In the previous example, the variables $y_1, y_2, y_3$ are
blocked, and $x_1$ and $x_2$ are not blocked. 
The set of vertices $\{y_1,y_2,y_3\}$ forms a strongly connected component,
which is not a sink component, because there are outgoing edges. (Moreover, the
variables in $K$ are blocked.) The singleton-set $\{x_1\}$ is a strongly
connected component without outgoing edges and without blocked vertices, and
thus $x_1$ is a sink.

The following lemma shows an important consequence of \lex-closure of 
constraints.
\begin{lemma}
\label{lem:component-contract}
Let $K$ be a sink component of the graph $G_\Phi$ for an instance $\Phi$ with \lex-closed constraints.  Then all variables from $K$ must have equal values in
all solutions of $\Phi$.
\end{lemma}

\begin{proof}
We assume that $\Phi$ has a solution, and that $K$ has at least two vertices 
(otherwise the lemma is trivial).
Let $t$ be a solution of $\Phi$, 
and let $M \subseteq K$ be the set of variables that have in $t$ 
the minimal value among the variables of the sink component $K$. 
If $M = K$, we are done. 

Otherwise, because $K$ is a strongly connected component, 
there is an edge in $G_\Phi$ from
some vertex $u \in M$ to some vertex $v \in K \setminus M$. 
By the definition of $G_\Phi$, 
there is a constraint $\phi$ in $\Phi$ such that 
whenever $u$ denotes the minimal value 
of a solution of $\phi$, then $v$ has to denote the minimal value as well.
By permuting arguments, we can assume without loss of generality
that $\phi$ is of the form $R(w_1,\dots,w_k)$ where $w_1=u$ and $w_2=v$.
Because $K$ is a sink component, the variable $u$ cannot be blocked,
and hence there is a minimal min-set $S$ for the first entry in $R$.
Clearly, $S$ contains $2$, because $v$ is the second 
argument of $\phi$.

Note that $G_\Phi$ contains an edge from $u$ to $w_i$ for all
$i \in S$. Since $K$ is a strongly connected component, 
all these variables $w_i$ are in $K$. Because
$u$ has in $t$ the minimal value among the variables in $K$,
there is no variable $w_i$, $i \in S$, which has a smaller
value than $u$ in $t$. This contradicts Lemma~\ref{lem:lex-free-set-equal},
because the value for $u$ in $t$ is different
than the value for $v$.
\end{proof}

Lemma~\ref{lem:component-contract} immediately implies that we can add
constraints of the type $x=y$ for all variables $x,y$ from the same sink 
component $K$. Equivalently, we can consider the CSP instance $\Phi'$ where
all the variables in $K$ are \emph{contracted}, i.e., where
all variables from $K$ are replaced by the same variable.
In some cases, a solution to a projected instance with ll-closed constraints
can be used to construct a solution to the original constraint.
We say that a tuple (in particular, a solution of an instance) $\overline x$ is
\emph{injective} if $x_i \neq x_j$ for all $i \neq j$. 

\begin{lemma}
\label{lem:injective-extension}
Let $\Phi$ be an instance of the CSP with variables $X$ and ll-closed
constraints. Let $x$ be a sink in $G_\Phi$. If the projection $\Phi'$ of
$\Phi$ to $X \setminus \{x\}$ has an injective solution, then $\Phi$ has an
injective solution as well.
\end{lemma}

\begin{proof}
Let $\overline s$ be an injective solution to $\Phi'$. 
Consider a constraint $\phi=R(x_1,\dots,x_k)$ from $\Phi$ 
that is imposed on $x$. 
By the definition of $\Phi'$ there is a tuple $t \in R$
such that $t$ agrees with $\overline s$ 
on $\{x_1,\dots,x_k\} \setminus\{x\}$.
Because $x$ is a sink, there is tuple
$t'\in R$ such that the entry corresponding to $x$ is strictly minimal. 
It is now easy to check that there
are automorphisms $\alpha, \beta$ of $(\mathbb Q,<)$ 
such that the tuple $t''=\alpha(ll(\beta(t'),
t))$ agrees with $\overline s$ on $X \setminus \{x\}$, and such that the 
entry corresponding to $x$ is strictly minimal. 
As $R$ is ll-closed, $t''\in R$. Thus we see
that for each constraint $ R(x_1,\dots,x_k)$ imposed on $x$ 
there is a tuple in $R$ where the entry corresponding
to $x$ is strictly minimal, and the rest of the tuple agrees with $\overline s$
on $X\setminus\{x\}$. Hence, we can extend $\overline s$ by assigning
to $x$ a value smaller than any value used in $\overline s$,
and the lemma readily follows.
\end{proof}

\ignore{
\begin{lemma}
\label{lem:reduction-final}
Let $\Phi$ be an instance of the CSP with ll-closed constraints. If all the
variables $X$ of $\Phi$ are sinks in $G_{\Phi}$, then any injective $|X|$-tuple
is a solution to $\Phi$.
\end{lemma}
\begin{proof}[Proof sketch]
The proof proceeds by induction. Let $\overline s$ a an injective tuple of
arity $|X|$.  If $\Phi$ has just one variable, the statement is trivially true.
Otherwise pick a variable $x_i \in X$ such that $s_i$ is minimal and consider
the projection $\Phi'$ of $\Phi$ to $X\setminus\{x_i\}$. By induction
$(s_1,\dots,s_{i-1},s_{i+1},\dots,s_{|X|})$ is a solution to $\Phi'$.  Now
Lemma~\ref{lem:injective-extension} shows that $\overline s$ is a solution to
$\Phi$. 
\end{proof}}

We are ready to state our algorithm for instances 
with ll-closed constraints; the algorithm works for both representations of the constraints (sets of weak linear orders, ll-Horn formulas).

\begin{figure}
\begin{center}
\small
\begin{minipage}[t]{7.7cm}
\begin{algorithm}\
\obeyspaces\obeylines\tt
Spec($\Phi$) $\{$
{\rm // Input: $\Phi$ constraints with variables $X$} 
{\rm // Output: If algorithm returns \false}
{\rm //   then $\Phi$ has no solution}
{\rm //   If $\Phi$ has an injective solution, then return \true}
{\rm //   Otherwise return $S \subseteq X$, $|S| \geq 2$, s.t. for all}
{\rm //   $x,y \in S$ we have $x=y$ in all solutions of $\Phi$}
\  $G :=$ ConstructGraph($\Phi$)
\  $Y := \emptyset$, $\Phi' := \Phi$, $G' := G$
\  While $G'$ contains a sink $s$
\    $Y := Y \cup \{s\}$
\    $\Phi' :=$ projection of $\Phi'$ to $X \setminus Y$
\    $G' :=$ ReconstructGraph($\Phi'$)
\  If $Y=X$ then return \true
\  else if $G'$ has sink component $S$
\    return $S$
\  else return \false
\  end if $\}$
\end{algorithm}
\end{minipage} \hspace{.3cm}
\begin{minipage}[t]{6.6cm}
\begin{algorithm}\label{alg:main} \
\obeyspaces\obeylines\tt
Solve($\Phi$): $\{$
{\rm // Input: instance $\Phi$ with variables X}
{\rm // Output: \true or \false}
\  $S := \text{Spec($\Phi$)}$
\  If $S =$ \false then return \false
\  else if $S =$ \true then return \true
\  else 
\    Let $\Phi'$ be contraction of $S$ in $\Phi$
\    return Solve($\Phi'$)
\  end if $\}$
\end{algorithm}
\end{minipage}
\end{center}
\caption{An algorithm for ll-closed constraints.}
\end{figure}

\begin{theorem}~\label{thm:alg}
The procedure Solve$(\Phi)$ in Algorithm~\ref{alg:main} 
decides whether a given set of ll-closed constraints $\Phi$
(where the relations are either represented by sets of weak linear orders,
or by ll-Horn formulas) has a solution. There is an implementation of the algorithm that runs 
in time $O(nm)$, where $n$ is the number of variables
of $\Phi$ and $m$ is the size of the input.
\end{theorem}

\begin{proof}
The correctness of the procedure Spec immediately implies the correctness 
of the procedure Solve.
In the procedure Spec, after iterated deletion of sinks in $G'$, 
we have to distinguish three cases. 

In the first case, $Y=X$. We prove by induction 
that $\Phi$ has an injective solution.
Let $x_1,\dots,x_n$ be the elements from $Y$ in the reverse 
order in which they were
included into $Y$.
For $0 \leq i \leq n$, let $\Phi_i$ be the instance $\Phi$ projected 
to $X \setminus \{x_1,\dots,x_i\}$. Note that $\Phi_0 = \Phi$,
and that $\Phi_n = \Phi'$ is the projection of $\Phi$ to the empty set,
which trivially has an injective solution.
We inductively assume that $\Phi_i$, for $i \leq n$, has an injective solution.
Then Lemma~\ref{lem:injective-extension} applied to $x_i$,
the instance $\Phi_{i-1}$, and the injective solution
to $\Phi_{i}$ implies that also $\Phi_{i-1}$ has an injective solution.
By induction, $\Phi_i$ has an injective solution for all $0 \leq i \leq n$, 
and in particular $\Phi_0=\Phi$ has an injective solution.
Therefore, the output \true\ of Spec is correct.

Otherwise, 
in the second case, $G'$ contains a sink component $S$ with $|S| \geq 2$. 
We claim that for all
variables $x,y \in S$ we have $x=y$ in all solutions to $\Phi$.
Lemma~\ref{lem:component-contract}
applied to the projection of $\Phi$ to $X \setminus Y$ 
implies that whenever some variables are in the same sink component, 
they must have the same value in every
solution, and hence the output is correct in this case as well.

In the third case, 
$Y \neq X$, but $G'$ does not contain a sink component.
Note that in every solution to $\Phi'$ some variable must take
the minimal value. However, since each strongly connected component
without outgoing edges contains a blocked vertex, there is no variable
that can denote the minimal element, and hence $\Phi'$ has no solution.
Because $\Phi'$ is a projection of $\Phi$ to $X \setminus Y$,
the instance $\Phi$ is inconsistent as well.

Since in each recursive call of Solve the instance
in the argument has at least one variable less, Solve is executed at most $n$
times. It is not difficult to implement the algorithm such that
the total running time is cubic in the input size.
However, it is possible to implicitely represent the constraint 
graph and to implement all sub-procedures
such that the total running time is in $O(nm)$,
for both types of representations of the constraints 
studied in this paper. 
We will now describe the details how this can be achieved.

We have already shown the correctness of the algorithm,
and only have to discuss how to implement the algorithm
such that it runs in $O(nm)$. In fact, we describe an
implementation of the procedure Spec that is linear in the input size.

First we show how to deal with constraints represented by ll-Horn clauses.
Observe that if an ll-Horn clause 
has a non-empty left hand side of the implication,
then a constraint for this clause creates
neither edges nor blocked vertices in the constraint graph. 
Also constraints of the type $z_0 > z_1 \vee \dots \vee z_0 > z_l$ 
do not create edges in $G_\Phi$.
Thus, when
constructing $G_\Phi$, we only care about constraints of the type
$z_0 > z_1 \vee \dots \vee z_0 > z_l \vee (z_0=z_1=\dots=z_l)$.
For such constraints we add edges from $z_0$ to $z_1,\dots,z_l$ 
to $G_\Phi$.

When the constrains in the input instance are represented by sets of weak linear orders we have to be more careful, and do not represent the edges of $G_\Phi$ explicitly, since there might be quadratically many edges which spoils the desired running time.  We sort the
weak linear orders $\prec$ in each set according to the number of equivalence classes (of the equivalence relation defined by $x \prec y \wedge y \prec x$).  Now, the
data structure contains for each variable and each constraint that is imposed
on this variable a reference to the weak linear order $\prec$ in this constraint such that $v$
is smallest with respect to $\prec$, and $\prec$ has the largest number of equivalence
classes.  
Moreover, for each element in each weak linear order $\prec$ 
we create a list that contains the elements from the same equivalence
class in $\prec$.  Finally, for each variable
$v$ we also have a list that contains the constraints that are imposed on $v$
and that block $v$.  With bucket sort, the total cost to set up this data structure
is linear in the input size.  Even though the constraint graph $G_\Phi$ is not
explicitely represented, it is possible to use the above data structure to
compute the strongly connected components of $G_\Phi$ in linear time, using
depth-first search. 

Now we have to describe how the algorithm finds sinks, how the data structure
is updated after projections, and how the algorithm finds sink components if
there is no sink left and not all variables have been projected out.  To find sinks and sink components, we also have to be able to determine efficiently
whether a node is blocked or not.

Initially, because we have computed the strongly connected components, and
because we know which variables are blocked, we can create a list that contains
all sinks of the initial instance.  Suppose that $s$ is a sink of $G$ at some
iteration of the while-loop. We then first compute the projection of $\Phi'$ to
$X \setminus Y$ by updating only the constraints imposed on $s$ in $\Phi'$.  At
this step we can also determine whether a constraint no longer blocks a
variable $v$, and in this case we can update the list of blocking constraints
for $v$. As soon as this list becomes empty, we know that $v$ is no longer
blocked. In this case, if $v$ does not have outgoing edges in the current
constraint graph, which we can determine efficiently using our updated data
structure, we add $v$ to the list of sinks.  The total number of operations we
have to perform in all iterations of the while-loop is then bounded by $m$.

Finally, if there is no sink left, but not all variables have been
projected out, then we can compute the strongly connected
components of the resulting constraint (again, this can be
done in linear time using depth-first search on our data structure), 
and since we know which variables are blocked, we can also find the
sink components.


Note that we can assume that $n$ is smaller than $m$.
Otherwise, the constraint is not connected (we use the notion of connectivity 
for instances of the CSP 
as e.g.~in~\cite{HNBook}). 
We can in this case use the same implementation, 
analyse the running time for each of the connected components separately, 
and get the same result.

This concludes the proof 
that for both representations studied in this paper
the algorithm can be implemented such that
it runs in time $O(nm)$.
\end{proof}

%% file: unbounded.tex
\section{\X-closed Constraints and Datalog}
In this section, we prove that the constraint
satisfaction problem for ll-closed constraints cannot be solved by Datalog
programs\footnote{This result should not be confused with the weaker fact
that establishing $k$-consistency does not imply global consistency, for any
$k$. This was shown for Ord-Horn in~\cite{Koubarakis}.  But recall that
Ord-Horn \emph{can} be solved by a Datalog program~\cite{Nebel}.}.
%
%
For simplicity, the definition of the sematics of Datalog that we use here
will be purely operational; 
for the standard semantical approach
to the evaluation of Datalog programs see~\cite{EbbinghausFlum}.
A Datalog program is a finite set of Horn clauses, 
i.e., clauses of the 
form $\psi \leftarrow \phi_1, \dots, \phi_l$, where $l \geq 0$
and where $\psi,\phi_1,\dots,\phi_l$ are atomic formulas of the form
$R(\overline x)$. The formula $\psi$
is called the \emph{head} of the rule, and $\phi_1,\dots,\phi_l$ are called the \emph{body}. We assume that all variables in the
head also occur in the body.
The relation symbols occurring in the head of some clause are called
\emph{intentional}, and all other relation symbols in the clauses are 
called \emph{extensional}.

If $\Gamma$ is a finite \tcl, 
we might use Datalog programs to solve CSP$(\Gamma)$ as follows.
Let $\Pi$ be a Datalog program whose extensional symbols are from $\Gamma$.
We assume that there is one distinguished 0-ary intentional relation symbol
\false. Now, suppose we are given an instance $\Phi$ of CSP$(\Gamma)$.
An \emph{evaluation} of $\Pi$ on $\Phi$ proceeds in steps $i=0,1,\dots$
At each step $i$ we maintain a set of literals $\Phi^i$ with extensional and intentional relation symbols; it always holds that $\Phi^{i} \subset \Phi^{i+1}$.
Each clause of $\Pi$ is understood as a rule that may derive a new literal
from the literals in $\Phi^i$.
Initially, we have $\Phi^0 := \Phi$. Now suppose that 
$R_1(x^1_1,\dots,x^1_{k_1}), \dots, R_l(x^l_1,\dots,x^l_{k_l})$
are literals in $\Phi^i$, and 
$R_0(y_1^0,\dots,y_{k_0}^0) \leftarrow R_1(y_1^1,\dots,y_{k_1}^1),\dots,
R_l(y^l_1,\dots,y^l_{k_l})$ 
is a rule from $\Pi$, 
where $y_{j}^i = y_{j'}^{i'}$ if and only if $x_j^i = x_{j'}^{i'}$.  
Then $R_0(x^0_1,\dots,x^0_l)$ is the newly derived literal in $\Phi^{i+1}$,
where $x^0_j = x^i_{j'}$ if and only if $y^0_j = y^i_{j'}$.
The procedure stops if no new literal can be derived.
We say that $\Pi$ solves CSP$(\Gamma)$, if for every instance
$\Phi$ of CSP$(\Gamma)$ there exists an evaluation of $\Pi$ on 
$\Phi$ that derives \false\ if and only if $\Phi$ has no solution.

We want to remark that the so-called method of \emph{establishing path-consistency}, which is very well-known and frequently applied in Artificial Intelligence, can be formulated with Datalog programs where the intentional 
symbols are at most binary and all rules use at most three variables 
in the body.

We prove that already for the \tcl\ 
that only consists of $R^{\min}$ there is no Datalog program that
solves the corresponding constraint satisfaction problem.
We use a pebble-game characterization
of the expressive power of Datalog, which was originally shown in~\cite{FederVardi} and~\cite{KolaitisVardi} for finite domain constraint satisfaction, and which holds for a wide variety of infinite domain 
constraint languages as well, including qualitative \tcl s\  Ê
(see the journal version of~\cite{BodDalJournal}).

Let $\Gamma$ be a finite \tcl, and let
$\Phi$ be an instance of CSP$(\Gamma)$.
Then the \emph{existential $k$-pebble game on $\Phi$} 
is the following game between the players 
\emph{Spoiler} and \emph{Duplicator}.
Spoiler has $k$ pebbles $p_1, \dots, p_k$. 
He places his pebbles on variables from $\Phi$.
Initially, no pebbles are placed. In each round of the game
Spoiler picks some of these pebbles.
If they are already placed on $\Phi$, 
then Spoiler first removes them from $\Phi$.
He then places the pebbles on variables from $\Phi$, 
and Duplicator responds by assigning elements from $\mathbb Q$ to
these variables. This assignment has to satisfy all the constraints $\phi \in \Phi$ where all variables in $\phi$ are pebbled, otherwise
Spoiler wins the game. 
Duplicator wins, if the game continues \emph{forever}, i.e., 
if Spoiler can never win the game. 

\begin{theorem}[from \cite{BodDal}]\label{thm:pebblethm}
Let $\Gamma$ be a finite \tcl.
There is no Datalog program that solves CSP$(\Gamma)$ if and only
if for every $k$ there exists an inconsistent instance of CSP$(\Gamma)$ 
such that Duplicator wins the existential $k$-pebble game on $\Phi$.
\end{theorem}

The rest of this section is devoted to the proof of the following theorem.

\begin{theorem}\label{thm:unbounded}
There is no Datalog program that solves CSP$(\{R^{\min}\})$.
\end{theorem}

\begin{proof}
Let $k$ be an arbitrary number. To apply Theorem~\ref{thm:pebblethm}
we have to construct an inconsistent instance $\Phi$ of CSP$(\{R^{\min}\})$ 
such that Duplicator wins the existential $k$-pebble game on $\Phi$.

For this, let $G$ be a 4-regular graph of girth at least $2k+1$, i.e., all cycles in $G$
have more than $2k$ vertices.
It is known and easy to see that such graphs exist, e.g.\ with
the methods in~\cite{JLR}.
Orient the edges in $G$ such that there are exactly two outgoing and two
incoming edges for each vertex in $G$. Since $G$ is 4-regular, there exists 
an Euler tour for $G$ (see e.g.~\cite{Diestel}), 
which shows that such an orientation exists. 

Now we can define our instance $\Phi$
of CSP$(\{R^{\min}\})$ as follows. The variables of $\Phi$ are the vertices
from $G$. The instance $\Phi$ contains the constraint $R^{\min}(w,u,v)$
iff $uw$ and $vw$ are the two incoming edges at vertex $w$. We claim that
$\Phi$ does not have a solution: if there was a solution,
some variable $w$ must denote the minimal value. But for 
every variable $w$ we find a constraint $R^{\min}(w,u,v)$ in $\Phi$, and this
constraint is violated since either $u$ or $v$ must be strictly
smaller than $w$.

We now show that Duplicator has a winning strategy for the existential
$k$-pebble game on this instance.
Consider a connected non-empty subgraph $G'$ 
of $G$ having at most $2k$ vertices
where only one vertex $r$ has no outgoing edges,
and where all vertices have either two or no incoming edges.
Since $G$ has girth $2k+1$, $G'$ must be a binary tree with root $r$.
We call $G'$ \emph{dominated}, if all leaves in $G'$ are pebbled.

Duplicator always maintains the property that whenever 
the root $r$ in a dominated
tree is pebbled during the game, then 
the value assigned to $r$ is strictly larger than the minimum
of all the values assigned
to the leaves. 
Clearly, this property is satisfied at the beginning of the game.

Suppose that during the game Spoiler pebbles the variable $u$. 
Let $T_1,\dots,T_s$ be those newly created dominated trees in $G$
that have pebbled roots $r_1,\dots,r_s$, for $s\geq 0$.
If $s>0$, let $r_i$ be the root that received the minimal value $a$
among all the roots $r_1,\dots,r_s$.
We claim that if $u$ is the root of a dominated tree $T$, then
$a$ is strictly larger than the minimum $b$ of
all the values assigned
to the leaves of $T$. Otherwise, the graph $T \cup T_i$ was a dominated
tree (since the number of pebbles is at most $k$) that violates the invariant even before the variable $u$ has
been pebbled, a contradiction.
Therefore, in this case Duplicator can choose a value $c$ between $b$ and $a$
for the variable $u$. Since $c$ is smaller than $a$, 
in all the new dominated trees 
$T_1,\dots,T_s$ in $G$ the value assigned to $r_1,\dots,r_s$ is
strictly larger than $c$, and hence the invariant is preserved.
In particular, if $R^{min}(w,u,v)$ (or $R^{min}(w,v,u)$) is 
a constraint in $\Phi$ where $w$ and $v$
have been pebbled, then this constraint is satisfied by the assignment.

Since $c$ is larger than $b$,
this choice also guarantees that if $v,v'$ are pebbled variables
then any constraint of the form $R^{\min}(u,v,v')$
is satisfied, because in this case the variables $u,v,v'$ 
induce a dominated tree with root $u$ in $G$. 

If there is no dominated tree $T$ where $u$ is the root, then
Duplicator assigns a value to $u$ that is smaller than all values
assigned to other variables.
If $s=0$, Duplicator plays a value that is larger than all values
assigned to other variables.
In both cases it is easy to check that Duplicator maintains
the invariant, and satisfies all constraints $\phi \in \Phi$ where
all variables are pebbled.
By induction, we have shown that Duplicator has a winning strategy
for the existential $k$-pebble game on $\Phi$.
\end{proof}

%% file: outlook.tex
\section{Conclusion}
While most of the polynomial algorithms that are known
and used to solve infinite-domain constraint satisfaction problems are based on
local consistency techniques, we used graph algorithms on an appropriately
defined notion of constraint graph to both improve applicability (the
constraint languages we can solve with our approach contain constraint
relations whose CSP can not be solved with local consistency techniques --
Theorem~\ref{thm:unbounded}) and running time (our algorithm has quadratic
running time, whereas resolution or establishing path consistency would require
cubic time).  We believe that similar approaches can lead to faster algorithms
and larger tractable languages for many problems where the only known
algorithms are based on local consistency techniques.